\pdfoutput=1

\documentclass[11pt]{article}

\usepackage{acl}

\usepackage{times}
\usepackage{latexsym}
\usepackage{microtype}
\usepackage{graphicx}
\usepackage{adjustbox}
\usepackage{multirow}
\usepackage{booktabs}
\usepackage{tabularx}
\usepackage[inline]{enumitem}
\usepackage{flushend}
\usepackage{balance}
\usepackage[T1]{fontenc}

\usepackage[utf8]{inputenc}

\usepackage[bottom]{footmisc}
\usepackage[section]{placeins}

\usepackage{amsmath}
\usepackage{csquotes}
\usepackage{subcaption}

\title{Scaled Prompt-Tuning for Few-Shot Natural Language Generation}

\author{Ting Hu \\
  Hasso Plattner Institute\\
  University of Potsdam \\
  Potsdam, Germany \\
  \texttt{ting.hu@hpi.de} \\
  \And
  Christoph Meinel\\
  Hasso Plattner Institute\\
  University of Potsdam \\
  Potsdam, Germany \\
  \texttt{meinel@hpi.de}\\
  \And
  Haojin Yang \\
  Hasso Plattner Institute\\
  University of Potsdam \\
  Potsdam, Germany \\
  \texttt{haojin.yang@hpi.de} \\
  }

\begin{document}
\maketitle

\begin{abstract}
The increasingly Large Language Models (LLMs) demonstrate stronger language understanding and generation capabilities, while the memory demand and computation cost of fine-tuning LLMs on downstream tasks are non-negligible. Besides, fine-tuning generally requires a certain amount of data from individual tasks whilst data collection cost is another issue to consider in real-world applications. In this work, we focus on Parameter-Efficient Fine-Tuning (PEFT) methods for few-shot Natural Language Generation (NLG), which freeze most parameters in LLMs and tune a small subset of parameters in few-shot cases so that memory footprint, training cost, and labeling cost are reduced while maintaining or even improving the performance. We propose a Scaled Prompt-Tuning (SPT) method which surpasses conventional PT with better performance and generalization ability but without an obvious increase in training cost. Further study on intermediate SPT suggests the superior transferability of SPT in few-shot scenarios, providing a recipe for data-deficient and computation-limited circumstances. Moreover, a comprehensive comparison of existing PEFT methods reveals that certain approaches exhibiting decent performance with modest training cost such as Prefix-Tuning in prior study could struggle in few-shot NLG tasks, especially on challenging datasets.
\end{abstract}

\section{Introduction}

With the emergence and development of Large Language Models (LLMs), fine-tuning already becomes the mainstream paradigm in Natural Language Processing regardless of the recently emergent powerful foundation models such as GPT-4 and the comcomitant prompt engineering. Though fine-tuning offers an effective means of transferring the pre-trained knowledge to downstream tasks and necessitates a small quantity of textual data in comparison to pre-training corpora, it still demands a significant amount of memory on device and entails substantial training cost. These facilitate the development of another research topic, Parameter-Efficient Fine-Tuning (PEFT), which freezes most of the parameters in LLMs and merely tunes a small portion of them, resulting in reduced memory footprint and computation expense with comparable performance to conventional fine-tuning. Furthermore, the emergence of in-context learning \cite{brown2020language} instills optimism for few-shot scenarios and draw considerable research interest in few-shot PEFT methods, dedicated to the least cost of adopting LLMs to data-scarce and resource-limited scenarios.

The core question of PEFT is which parameters in LLMs are to be tuned so that we could tune as fewer parameters as possible with the least performance drop on downstream tasks. Many approaches, such as Adapter \cite{houlsby2019parameter} and Prompt-Tuning \cite{lester2021power}, varying in trainable parameters and performance have been proposed and applied in various fields. Despite some recent research on few-shot PEFT for Natural Language Understanding (NLU) tasks, our comprehension of PEFT methods on Natural Language Generation (NLG) tasks, including Meaning Representation (MR)-to-text and Knowledge Graph (KG)-to-text generation, in few-shot cases is insufficient, which motivates us to delve into the details. We conclude our contributions below.
\begin{itemize}
    \item We put forward Scaled Prompt-Tuning (SPT) which drastically outcompetes conventional Prompt-Tuning with negligible extra trainable parameters.
    \item SPT demonstrates better transferability than fine-tuning in few-shot cases, which provides a recipe in resource-limited environments without extra labeling cost via intermediate SPT.
    \item The comprehensive comparison of existing PEFT methods manifests that approaches that perform decently when a sufficient number of data instances are available such as Prefix-Tuning could face hurdles in few-shot cases, especially on challenging datasets.
\end{itemize}

\section{Related work}

\begin{table*}[!th]
\small
\begin{center}
    \begin{tabular}{c|p{3cm}p{3cm}p{2cm}}
    \hline
       Method & Base model & Tasks & Few-shot tasks\\
       \hline
       Adapter \cite{houlsby2019parameter} & BERT & GLUE & - \\
       \hline
       LoRA \cite{hu2021lora}  & \raggedright RoBERTa, DeBERTa, GPT-2 &  \raggedright GLUE, WikiSQL, SAMSum & RTE, MNLI \\
       \hline
       Compacter \cite{karimi2021compacter} & T5 & GLUE, SuperGLUE &  - \\
       \hline
       Unified \cite{he2021towards} & BART, RoBERTa & \raggedright XSum, WMT 2016, MNLI, SST-2 & - \\
       \hline
       IA3 \cite{liu2022few} & T0 & T0 tasks & RAFT \\
       \hline
       UniPELT \cite{mao2021unipelt} & BERT & GLUE & GLUE \\
       \hline
       Prompt-Tuning \cite{lester2021power} & T5 & GLUE, SuperGLUE & - \\
       \hline
       OptiPrompt \cite{zhong2021factual} & BERT & LAMA & - \\
       \hline
       WARP \cite{hambardzumyan2021warp} & RoBERTa & GLUE & FewGLUE \\
       \hline
       P-Tuning v2 \cite{liu2021p} & GPT-2, BERT & LAMA, SuperGLUE & - \\
       \hline
       Prefix-Tuning \cite{li2021prefix} & GPT-2, BART & NLG, XSum & FewGLUE \\
     \hline
    \end{tabular}
\end{center}
\caption{ Comparison of PEFT methods. SuperGLUE is a stickier benchmark originating from GLUE. FewGLUE is a benchmark for few-shot SuperGLUE tasks. RAFT is a real-world few-shot text classification benchmark. SAMSum and XSum are text summrization tasks. LAMA is a probe for analyzing the factual and commonsense knowledge contained in LLMs. WMT 2016 is a machine translation task. 
\label{pet_relatedwork} }
\end{table*}

PEFT approaches strive to tune a fraction of parameters and freeze most of the parameters in LLMs which demands less memory footprint and energy consumption. These methods differ in which parameters are tuned on downstream tasks. The very first PEFT work is Adapter \cite{houlsby2019parameter}, where small trainable bottleneck modules are inserted in BERT \cite{devlin2018bert} layers per task. More specific, the adapters are inserted at two positions of each layer: after the projection following Multi-Head Attention module and after two Feed-Forward Networks (FFNs). By tuning task-specific Adapters on GLUE benchmark \cite{wang2018glue}, the performance drop is within 0.4\% of that of fine-tuning, while only 3.6\% of the parameters are added. Intuitively, the insertion positions of the Adapters and the possibility of multi-task adapters are interesting questions. \citet{he2021towards} further demonstrate that adding adapters after the FFNs is sufficient to encapsulate and refine the task-specific information from the frozen parameters, since FFNs can better utilize modification at larger capacities. This effectively reduces 50\% of inserted adapters in previous work. On the other hand, AdapterFusion \cite{pfeiffer2020adapterfusion} conducts a two-stage learning algorithm: task-specific adapters training and combining adapters in the fusion module. The authors show that combining the knowledge from different tasks obtained by their corresponding adapters could be beneficial for each individual task. Despite their less memory demands and tuning costs, adapters actually introduce around 4-6\% extra inference time, since all parameters of BERT and inserted adapters are involved in inference. AdapterDrop \cite{ruckle2020adapterdrop} further proposes to drop a variable number of adapters from lower BERT layers. It dynamically reduces the computational overhead at run-time when performing inference over multiple tasks and maintains task performance to a large extent. Considering the Fully Connected (FC) layers in the bottleneck adapters still have a relatively large number of parameters, Compacter \cite{karimi2021compacter} introduces more efficient Parameterized Hypercomplex Multiplication (PHM) Layers to replace the FC layers, where the weight of each FC layer is computed as the sum of several Kronecker products. \citet{karimi2021compacter} further reduce the trainable parameters by letting the adapters across layers share slow weights and vary in fast rank-one matrices, resulting in Compacter++. Compacter works on par with fine-tuning when applied to T5-Base \cite{raffel2020exploring} on GLUE benchmark by only training 0.047\% of the parameters.

Another line of work insert extra trainable parameters in other formats instead of the bottleneck modules, including Prompt-Tuning and Prefix-Tuning. Prompt-Tuning is different from prompting, i.e., in-context learning. In-context learning blooms as the emergence of GPT-3 \cite{brown2020language}, which could directly adapt to some downstream tasks by inputting prompts instead of tuning parameters. The prompts usually have some descriptions of the tasks followed by several exemplars and specific content that we want the model to help with. Prompt-Tuning \cite{lester2021power} prepends a trainable soft prompt to the input embeddings of the model for specific downstream tasks. Only the continuous prompts are updated during training. This methods performs well when applied on models that are pretrained in a multi-task setting such as T5. However, the impact of the soft prompt could become weaker and weaker as a model goes deeper, then another line of work that more effectively modify the representations in the models show up. Prefix-Tuning \cite{li2021prefix} prepends a trainable continuous prefix to the input of each layer for an decoder-style model and two prefixes for the encoder-decoder model, separately. By learning only 0.1\% of the parameters of BART \cite{lewis2019bart}, Prefix-Tuning obtains comparable performance to fine-tuning on table-to-text generation tasks. P-Tuning v2 \cite{liu2021p} shares a similar idea to Prefix-Tuning, which inserts multi-layer prompts in the model while studying the performance on NLU tasks.

Other methods explore tuning other parameters in LLMs. Intrinsic SAID \cite{aghajanyan2020intrinsic} empirically shows that LLMs have very low intrinsic dimensions and proposes to tune the parameters in a lower-dimension subspace, which is achieved by a random linear projection via Fastfood transform. FISH Mask \cite{sung2021training}] selects a subset of parameters to update based on their estimated Fisher information. BitFit \cite{zaken2021bitfit} demonstrates that solely tuning the bias terms in BERT is competitive with fine-tuning with small-to-medium training data scale. It raises the hypothesis that fine-tuning is mainly about exposing knowledge induced by language-modeling training, rather than learning new task-specific linguistic knowledge. LoRA \cite{hu2021lora} injects trainable rank decomposition matrices into each Transformer layer based on the hypothesis that the change in weights during model adaptation has a low intrinsic rank. It does not introduce inference latency and reduce input sequence length while retaining high performance on RoBERTa, GPT-2 and even GPT-3. $\mathrm{(IA)}^{3}$~\cite{liu2022few} introduces three learned vectors to rescale the Keys and Values in the attention modules and the inner activations of the point-wise FFNs, respectively, via element-wise multiplication. This approach is conducted on T0 \cite{sanh2021multitask} which is pretrained in a multi-prompt and multi-task manner. UniPELT \cite{mao2021unipelt} further incorporates different PEFT methods as sub-modules, including Adapters, LoRA, and Prefix-Tuning, and learns to activate the ones that suit the current data or task setup the best via a gating mechanism on GLUE tasks.
Similarly, \citet{he2021towards} breaks down the design of various PEFT methods and re-frame them as modifications to specific hidden states in LLMs. It establishes a unified framework with Adapters, Prefix-Tuning, and LoRA in BART, and conducts experiments on NLU, summarization and machine translation tasks. 

We compare above methods from different aspects in Tab.~\ref{pet_relatedwork}. As we can see, there are few existing PEFT approaches centering on NLG tasks, not to mention few-shot NLG tasks. Therefore, the understanding and comprehension of the characteristics of PEFT methods on NLG tasks in few-shot cases is deficient, which we focus on in this work. Moreover, there are some related work proposing scaling related ideas which we compare with below. $\mathrm{(IA)}^{3}$~\cite{liu2022few} introduces three learned vectors to rescale the Keys, Values, and inner activations. \citet{he2021towards} propose scaled parallel adapters and demonstrate the scaling factor is significant for parallel adapters. Our proposed Scaled Prompt-Tuning could be regarded as a simplified variant of them which does not pose any extra modification to each Transformer layer and tune much less parameters than theirs.

\section{Method}

Conventional Prompt-Tuning freezes the parameters of LLMs and solely tunes the embeddings of $k$ additional tokens, i.e., soft prompt, for individual downstream tasks. Assume the input sequence with $l$ tokens is $X=\{x_{1}, x_{2}, \dots, x_{l}\}$. After going through the embedding layer, the sequence is represented by a matrix $X_{e} \in R^{l \times n_{e}}$, where $n_{e}$ is the hidden dimension of the embeddings. The trainable soft prompt, denoted as $X_{p} \in R^{k \times n_{e}}$, is prepended to the input sequence embedding $X_{e}$, resulting in the update embedding matrix $X_{h}=[X_{p};X_{e}] \in R^{(k+l) \times n_{e}}$, which is fed into the later blocks of LLMs for further computation, while only $X_{p}$ is optimized during training.

\begin{figure}[!th]
    \centering
    \includegraphics[width=0.45\textwidth]{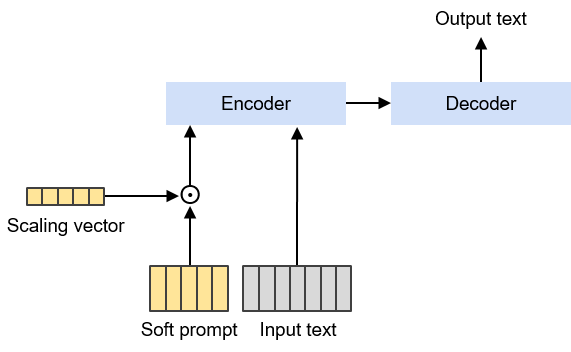}
\caption{The proposed Scaled Prompt-Tuning. The parameters of the encoder and decoder are frozen, while the soft prompt and scaling vector are trained for downstream tasks. \label{spt_arch}}
\end{figure}

We propose Scaled Prompt-Tuning where the trainable soft prompt $X_{p}$ with an additional trainable scaling vector $s \in R^{k \times 1}$ are employed. Consequently, the updated embedding matrix is $X_{h}=[s \odot X_{p};X_{e}]$, where $\odot$ denotes Hadamard product. That is, each scaling factor in the scaling vector $s$ is applied to scale the embedding of an individual soft token, yielding narrowed representation gap between the soft prompt and input sequence embeddings. Other alternatives could be using a single-value scaling factor or a scaling matrix $s \in R^{k \times n_{e}}$, while both of them empirically underperform the proposed one-dimension scaling vector. Fig.~\ref{spt_arch} depicts the proposed SPT method, and the encoder-decoder LLM we work on is T5-large \cite{raffel2020exploring} with a total 770M parameters.

\section{Experiments}

\subsection{Data and evaluation metrics}

Experiments are conducted on three NLG datasets: WebNLG 2020\footnote{\url{https://synalp.gitlabpages.inria.fr/webnlg-challenge/challenge_2020}}, E2E\footnote{\url{https://github.com/tuetschek/e2e-metrics}}, and DART\footnote{\url{https://github.com/Yale-LILY/dart}}. Since they contain a large number of instances, we sample subsets of instances from each dataset for few-shot tuning. The sampling process is implemented three times for each few-shot scenario.
WebNLG 2020 has 16 categories of KGs, and we sample from each category. E2E dataset has no definition of category, then we differentiate the instances based on the number of slot-value pairs in MRs, thereby categorizing them into 6 distinct groups and sampling from each group. The samples in DART are from 6 sources and we sample from each source.

The transformation of structured data into the inputs of LLMs is significant. This is related to the pre-training paradigm of the LLM that PEFT methods are applied to. In this work, the model we adopt is T5, pre-trained in a text-to-text manner. Therefore, we employ the structured data transformation method in Tab.~\ref{data_trans}. For WebNLG 2020 and DART, we linearize the triples in KGs and prepend token \textless S\textgreater, \textless P\textgreater, and \textless O\textgreater ~ to the subject, predicate, and object of each triple, respectively. Regarding E2E, we linearize the slot-value pairs in MRs, and insert token \textless S\textgreater~and \textless V\textgreater~before the slot and value, separately. This transformation process is demonstrated to be effective according to prior work \cite{kale2020text} and our experiments. The token  \textless S\textgreater, \textless P\textgreater, \textless O\textgreater, and \textless V\textgreater~are delimiters that maintain the structural information to some degree in the transformed text sequences.

\begin{table*}[!th]
\small
\begin{center}
\renewcommand{\arraystretch}{1.1}
    \begin{tabular}{p{14.5cm}}
     \hline
     E2E \\
    \hline
     \textbf{MR:} name[Aromi], eatType[coffee shop], food[French], customer rating[low], area[city centre], familyFridenly[no] \\
      \textbf{Transformed MR:} <S> name <V> Aromi <S> eatType <V> coffee shop <S> food <V> French <S> customer rating <V> low <S> area <V> city centre <S> familyFridenly <V> no\\
      \textbf{Reference:} In the city centre lies Aromi, a French coffee shop for adults with a low customer rating. \\
    \hline
    DART \\
    \hline
     \textbf{KG:} (David Davie Shelby, born/died, 1847-1914), (David Davie Shelby, active, 1899-1914), (David Davie Shelby, state, AL)\\
     \textbf{Transformed KG:} <S> David Davie Shelby <P> born/died <O> 1847-1914 <S> David Davie Shelby <P> active <O> 1899-1914 <S> David Davie Shelby <P> state <O> AL \\
     \textbf{Reference:} Judge David Davie Shelby from state AL was born in 1847 and died in 1914 , and is active during 1899 and 1914. \\
    \hline
    \end{tabular}
\end{center}
\caption{\label{data_trans} Examples of structured data transformation on E2E and DART. The transformed sequence and reference are used as the input of the encoder and ground truth output of the decoder for PEFT.}
\end{table*}

For each dataset, we employ the metrics provided in the benchmark for evaluation.
WebNLG 2020 applies BLEU, METEOR, chrF++, TER, BERTScore, and BLEURT. E2E uses BLEU, NIST, METEOR, ROUGE-L, and CIDEr. DART employs BLEU, METEOR, TER, BERTScore, MoverScore, and BLEURT. Most metrics measures the similarity between the generated sentences and the references from varied facets, the higher the better. TER measures how much entities in given structured data are correctly conveyed in generated sentences, the lower the better.
Considering the training instability of few-shot learning, we conduct experiments three times with distinct random seeds on each sampled subset. Eventually, we showcase the average evaluation results of nine experiments in each few-shot case below if not specially claimed.

\subsection{Approaches and implementation details}

\begin{table*}[!th]
\small
\begin{center}
    \begin{tabular}{c|ccc}
      \hline
      Method  & Learning rate & Config & \% trainable params  \\
      \hline
      Fine-Tuning & 1e-4 & - & 100.0 \\
      Prompt-Tuning &  5e-1 & Prompt length: 50 & 0.007\\
      SPT &  5e-1 &  Prompt length: 50 & 0.007 \\
      Adapter & 1e-4 &  $r$: 16 & 0.824 \\
      LoRA & 1e-4, 5e-4 & Rank: 8 & 0.306  \\
      Compacter & 3e-3 & PHM dim: 8, $r$: 16 & 0.053 \\
      Prefix-Tuning  & 5e-2, 1e-1 & Prefix length: 5, 10 & 0.096, 0.192 \\
      IA3 & 3e-3 & Rank: 1 & 0.045 \\
      UniPELT & 1e-4, 1e-3 & Rank: 8, $r$: 16,  Prefix length: 5, 10  & 1.194, 1.258 \\
     \hline
    \end{tabular}
\end{center}
\caption{Hyperparameter settings of PEFT methods. SPT represents the proposed Scaled Prompt-Tuning. The reduction factor shared by several methods is denoted as $r$. For methods that have two learning rates or two settings in the configuration, the former is shared on WebNLG 2020 and E2E dataset, and the latter is for DART dataset. Fine-Tuning trains all the parameters in the model, resulting in 100\% trainable parameters. Prompt-Tuning and SPT trains the least number of parameters.\label{pet_lr} }
\end{table*}

We compare with the proposed SPT with related work, including Adapter, LoRA, Compacter, Prefix-Tuning, Prompt-Tuning, and more recent methods, IA3 and UniPELT. These PEFT approaches are applied to T5-large with 770M parameters. For Prompt-Tuning and SPT, the length of the prompt is 50 on all datasets. Regarding Prefix-Tuning, we directly train the prefix parameters rather than replacing them by bottleneck modules to compare with our proposed method more closely. Moreover, bottleneck modules do not outperform simple prefix parameters in our experiments despite their effectiveness in related work \cite{li2021prefix}. For other methods, we start from the default settings provided in adapter-transformers library\footnote{\url{https://github.com/adapter-hub/adapter-transformers}} and further tune them on different datasets if the defaults are not applicable. DART is a challenging dataset and requires more tuning on learning rates and configurations. The detailed settings of PEFT approaches are summarized in Tab.~\ref{pet_lr}. For each experiment, we save the checkpoint resulting in the highest BLEU score on dev set and further evaluate it on test set.

\subsection{Comparison of few-shot PEFT methods}

\subsubsection{WebNLG 2020}

\begin{figure}[!th]
    \centering

    \begin{subfigure}[b]{0.45\textwidth}
    \includegraphics[width=\textwidth]{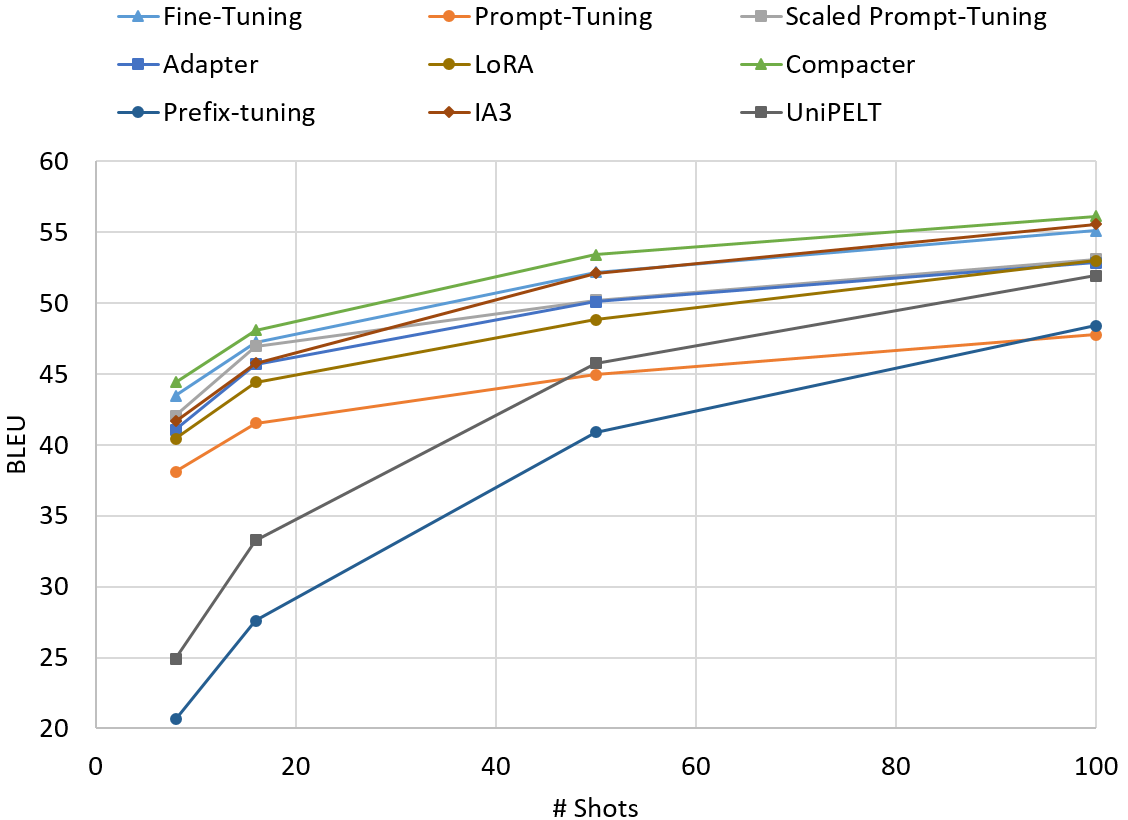}
    \caption{WebNLG 2020}
    \end{subfigure}

    \begin{subfigure}[b]{0.45\textwidth}
    \includegraphics[width=\textwidth]{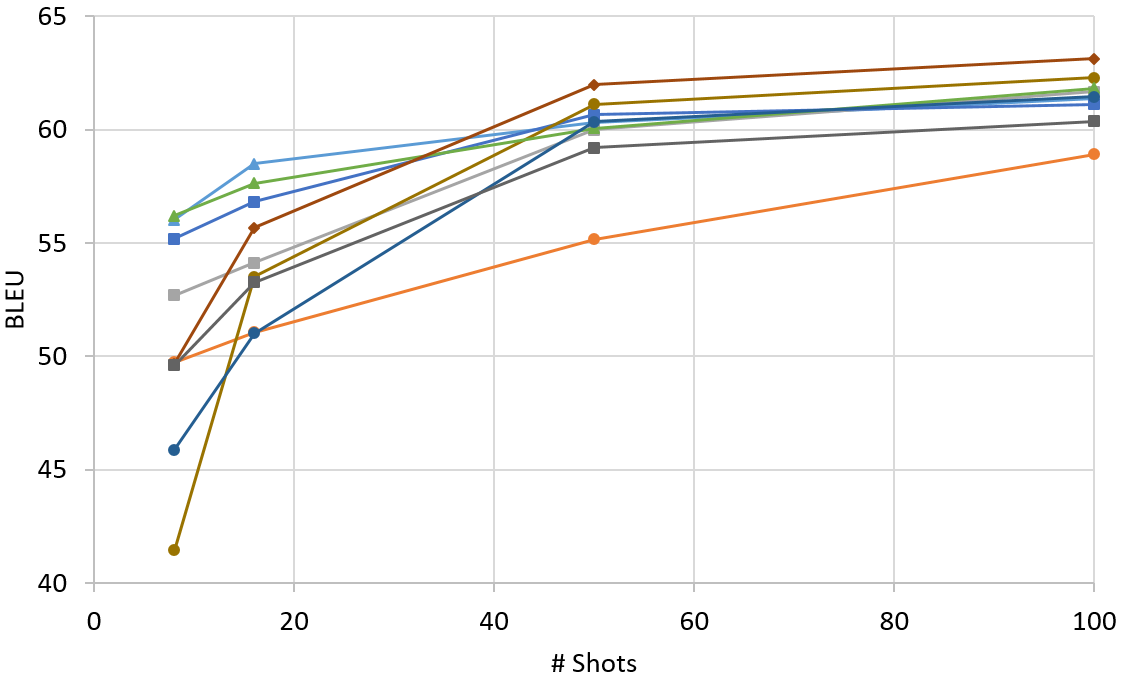}
    \caption{E2E}
    \end{subfigure}

    \begin{subfigure}[b]{0.45\textwidth}
    \includegraphics[width=\textwidth]{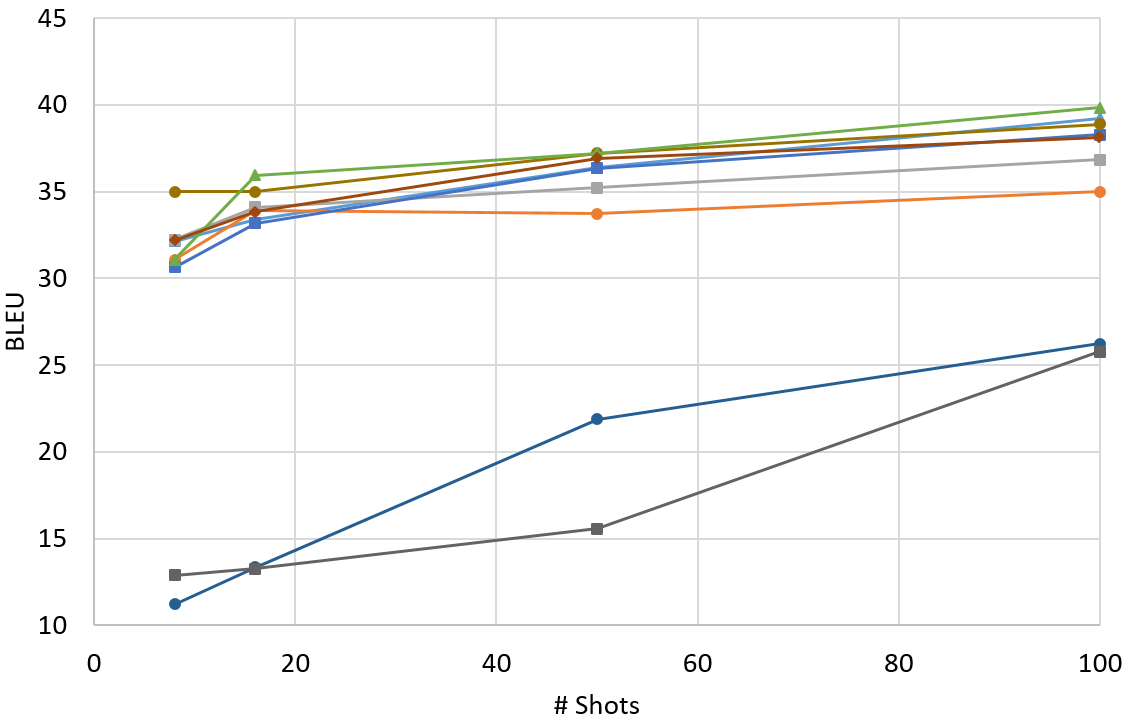}
    \caption{DART}
    \end{subfigure}

\caption{The performance of PEFT methods in few-shot cases on three datasets. \label{pet_webnlg2020}}
\end{figure}

We evaluate the PEFT methods in 8-, 16-, 50- and 100-shot cases on WebNLG 2020 dataset in Fig.~\ref{pet_webnlg2020} (a). Among the methods present, Compacter attains the best performance and even outperforms Fine-Tuning while only tuning 0.053\% of the parameters. Prefix-Tuning obviously falls behind others in 8-shot and 16-shot scenarios. With such low BLEU scores, the generated texts almost merely copy some of the given structured data and fail to form fluent sentences. Additionally, replacing prefix vectors with more complicated bottleneck modules and inserting prefixes in the decoder do not see obvious performance boost. These phenomena are not manifested in related work such as \citet{li2021prefix} since most of them conduct the whole-dataset tuning, which demonstrates the acceptable performance of Prefix-Tuning. In fact, Prefix-Tuning gradually becomes powerful as the number of available instances increases and even outperforms Prompt-Tuning in 100-shot case, showcasing the effectiveness of Prefix-Tuning. Therefore, our conclusion is that Prefix-Tuning is very sensitive to the number of training instances on WebNLG 2020 and performs poorly in extremely few-shot cases.

Another method that attracts our attention in Fig.~\ref{pet_webnlg2020} (a) is UniPELT, a combination of LoRA, Prefix-Tuning, and Adapter, which claims to take advantage of all of them. However, UniPELT drastically lags behind Adapter and LoRA while outperforms Prefix-Tuning. It seems that Prefix-Tuning plays a major role in UniPELT and thus leads to performance inferior to other methods. We analyze that three approaches involved in UniPELT have different convergence property and are sensitive to hyperparameters such as learning rate, as Tab.~\ref{pet_lr} shows. This disparity is amplified on generation tasks and an individual learning rate could not facilitate the convergence of all the trainable parameters in few-shot scenarios, yielding a subpar performance.

On the other hand, the proposed SPT obviously surpasses Prompt-Tuning and is on-par with Adapter. Tab.~\ref{petsome_webnlg2020} further displays the detailed evaluation results of Fine-Tuning, Prompt-Tuning, SPT, and the top-performing method in each few-shot case to shed light on the improvement of the proposed method and its performance gap to the best. The results of other metrics such as MRTEOR and BLEURT almost align with that of BLEU while the gap of the scores could be narrower. Therefore, we mainly refer to BLEU scores when discussing different approaches' performance below unless explicitly stated.

As we can see, SPT outcompetes conventional Prompt-Tuning by around 5.3 points in BLEU score under 16-, 50- and 100-shot circumstances. Compacter is the best-performing approach in all few-shot cases and surpasses SPT by a maximum of 3.2-point BLEU score. When the whole dataset is involved in tuning, UniPELT achieves superior performance on most evaluation metrics and even outperforms Fine-Tuning. This is consistent with \citet{mao2021unipelt} in that UniPELT combines several PEFT methods and achieves performance improvement on downstream datasets.

\begin{table}[!th]
\small
\setlength\tabcolsep{1.5pt}
\begin{center}
    \begin{tabular}{c|l|cccccc}
    \hline
      \# Shots & Method & BLEU & MET & chrF++ & TER & BS & BLEURT \\
     \hline
     \multirow{4}{*}{8} & FT & 43.5 & 0.37 & 0.61 & 0.52  & \textbf{0.95} & 0.33 \\
     & PT & 38.1 & 0.32 & 0.54 & 0.57 & 0.94 & 0.18  \\
     & SPT & 42.0 & 0.34 & 0.57 & 0.52 & 0.94 & 0.27  \\
     & Com & \textbf{44.4} & \textbf{0.38} & \textbf{0.62} & \textbf{0.51} & \textbf{0.95} & \textbf{0.35}  \\
     \hline
     \multirow{4}{*}{16} & FT & 47.2 & \textbf{0.39} & \textbf{0.65} & 0.47 &  \textbf{0.95} & \textbf{0.41}  \\
     & PT & 41.5 & 0.35 & 0.58 & 0.52 & 0.94 & 0.29 \\
     & SPT & 46.9 & 0.38 & 0.63 & \textbf{0.46} & \textbf{0.95} & 0.38  \\
     & Com & \textbf{48.1} & \textbf{0.39} & \textbf{0.65} & \textbf{0.46} & \textbf{0.95} & 0.40 \\
     \hline
     \multirow{4}{*}{50} & FT & 52.2 & 0.41 & 0.68 &  0.42 &  \textbf{0.96} &  0.46 \\
     & PT & 44.9 & 0.37 & 0.61 & 0.49 & 0.95 & 0.34 \\
     & SPT & 50.2 & 0.41 & 0.68 & 0.41 & \textbf{0.96} & 0.48  \\
     & Com & \textbf{53.4}  & \textbf{0.42} & \textbf{0.69} & \textbf{0.40} & \textbf{0.96} & \textbf{0.49}  \\
     \hline
     \multirow{4}{*}{100} & FT & 55.1 & 0.42 & 0.70 & 0.39 & 0.96  & 0.50 \\
     & PT & 47.8 & 0.39 & 0.65 & 0.45 & 0.96 & 0.41 \\
     & SPT & 53.1 & 0.42 & 0.69 & 0.39 & 0.96 & 0.50  \\
     & Com & \textbf{56.1} & \textbf{0.43} & \textbf{0.71} & \textbf{0.38} & 0.96 & \textbf{0.52}  \\
     \hline
     \multirow{4}{*}{All} & FT & 64.1 & 0.46 & 0.75 & 0.32 & 0.97 & 0.59 \\
     & PT & 57.5 & 0.43 & 0.71 & 0.35 & 0.97 & 0.55 \\
     & SPT & 59.2 & 0.44 & 0.72 & 0.34 & 0.97 & 0.56  \\
     & Uni & \textbf{65.8} & \textbf{0.47} & \textbf{0.76} & \textbf{0.30} & 0.97 & \textbf{0.61}  \\
     \hline
    \end{tabular}
\end{center}
\caption{ Comprehensive evaluation results of several PEFT methods on WebNLG 2020 in few-shot cases. FT denotes Fine-Tuning, PT represents Prompt-Tuning, and SPT denotes Scaled Prompt-Tuning. Com represents Compacter and Uni represents UniPELT. All means the whole dataset is involved in tuning. The top-performing method in each few-shot case is listed for comparison. BS represents BERTScore and MET represents METEOR. \label{petsome_webnlg2020} }
\end{table}

%
%

\subsubsection{E2E}

The tendency of PEFT methods' performance on E2E dataset in few-shot cases is different from that on WebNLG 2020, as Fig.~\ref{pet_webnlg2020} (b) depicts. No single method could dominate in all few-shot scenarios. Compacter stands out in 8- and 16-shot cases. IA3 then prevails over others when more samples are available for tuning.  Both Prefix-Tuning and UniPELT perform much better on E2E dataset than on WebNLG 2020. We conjecture the reason is that E2E is a less challenging dataset with simpler structured data, and tuning the prefix vectors is sufficient to refine the hidden representations for generation. Moreover, LoRA and Prompt-Tuning become the underachievers in few-shot cases according to Fig.~\ref{pet_webnlg2020} (b).


Furthermore, SPT still outperforms Prompt-Tuning with a large margin, where the details are listed in Tab.~\ref{petsome_e2e}. The largest performance gap between them is 4.8 points in BLEU score in 50-shot case. The largest discrepancy in performance between SPT and the top-performing method is 3.5-point BLEU score in 8-shot scenario. Similarly, UniPELT prevails over other approaches and is on-par with Fine-Tuning when all the samples in the dataset are used for tuning. Importantly, the proposed SPT merely falls behind UniPELT by 1.4 points in BLEU score while UniPELT tunes 170x as many parameters as SPT.

\begin{table}[!th]
\small
\setlength\tabcolsep{4.5pt}
\begin{center}
    \begin{tabular}{c|l|ccccc}
    \hline
      \# Shots & Method & BLEU & NIST & MET & R-L & CID \\
     \hline
     \multirow{4}{*}{8} & FT & 56.0 & 6.36 & \textbf{0.37} & \textbf{0.62} & 1.53 \\
     & PT & 49.7 & 5.60 & 0.34 & \textbf{0.62} & 1.39 \\
     & SPT & 52.7 & 5.64 & 0.34 & 0.60 & 1.34 \\
     & Com & \textbf{56.2} & \textbf{6.99} & \textbf{0.37} & \textbf{0.62} & \textbf{1.57} \\
     \hline
     \multirow{4}{*}{16} & FT & \textbf{58.5} & 7.24 & 0.37 & \textbf{0.63} & 1.76  \\
     & PT & 51.1 & 6.78 & 0.36 & 0.61 & 1.60 \\
     & SPT & 54.1 & 7.46 & 0.37 & 0.60 & 1.72 \\
     & Com & 57.6 & \textbf{7.49} & \textbf{0.39} & \textbf{0.63} & \textbf{1.81} \\
     \hline
     \multirow{4}{*}{50} & FT & 60.3 & 7.83 & 0.40 & 0.62 & 1.68 \\
     & PT & 55.2 & 6.56 & 0.36 & 0.61 & 1.60 \\
     & SPT & 60.0 & 7.49 & 0.38 & 0.62 & 1.80  \\
     & IA3 & \textbf{62.0} & \textbf{8.00} & \textbf{0.41} & \textbf{0.65} & \textbf{2.00} \\
     \hline
     \multirow{4}{*}{100} & FT & 61.4 & 7.77 & 0.39 & 0.65 & 1.81 \\
     & PT & 58.9 & 7.44 & 0.37 & 0.61 & 1.79 \\
     & SPT & 61.7 & 7.89	& \textbf{0.45} & \textbf{0.66} & 2.03  \\
     & IA3 & \textbf{63.1} & \textbf{8.09} & 0.43 & \textbf{0.66} & \textbf{2.08} \\
     \hline
     \multirow{4}{*}{All} & FT & 66.3 & \textbf{8.57} & 0.45 & 0.69 & 2.20 \\
     & PT & 64.6 & 8.29 & 0.45 & 0.67 & 2.27 \\
     & SPT & 65.0 & 8.39 & \textbf{0.46} & 0.68 & \textbf{2.31}  \\
     & Uni & \textbf{66.4} & 8.44 & \textbf{0.46} & \textbf{0.70} & 2.23 \\
     \hline
    \end{tabular}
\end{center}
\caption{ Comprehensive evaluation results of several PEFT methods on E2E in few-shot cases. FT denotes Fine-Tuning, PT represents Prompt-Tuning, and SPT denotes Scaled Prompt-Tuning. Com represents Compacter and Uni represents UniPELT. MET denotes METEOR, R-L represents ROUGE-L, and CID denotes CIDEr. All indicates the whole dataset is involved in tuning. The best-performing method in each few-shot case is listed for comparison. \label{petsome_e2e} }
\end{table}

\subsubsection{DART}

This dataset is the most challenging one among the datasets we work on, as the instances are from various domains and in distinct original formats. As Fig.~\ref{pet_webnlg2020} (c) displays, LoRA is the best-performing in 8-shot case, and Compacter surpasses others in 16-, 50-, and 100-shot scenarios. Prefix-tuning and UniPELT again drastically underperform others even in 100-shot case, implicitly reflecting that the instances in DART are more challenging than that in WebNLG 2020.


Moreover, SPT stably surpasses Prompt-Tuning, and the largest performance improvement is 1.8 points in BLEU score in 100-shot case, as Tab.~\ref{petsome_dart} elaborates. Meanwhile, SPT performs worse than Compacter by 3-point BLEU score. When the whole dataset is employed in tuning, the performance gap between SPT and Adapter, the current top-1 method, is 1.8-point BLEU score. The proposed SPT is again promising considering Adapter trains more than 100x the number of parameters of SPT.

\begin{table}[!th]
\small
\setlength\tabcolsep{2.0pt}
\begin{center}
    \begin{tabular}{c|l|cccccc}
    \hline
      \# Shots & Method & BLEU & MET  & TER & MS & BS & BLEURT\\
     \hline
     \multirow{4}{*}{8} & FT & 32.1 & \textbf{0.28} & 0.58 & 0.62 & 0.91 & \textbf{0.10} \\
     & PT & 31.1 & 0.22 & 0.64 & 0.60 & 0.91 & -0.10  \\
     & SPT & 32.2 &  \textbf{0.28}	& 0.58 & \textbf{0.63} & \textbf{0.92} & \textbf{0.10}  \\
     & LoRA & \textbf{35.0} &  \textbf{0.28}	& \textbf{0.59} & 0.62 & \textbf{0.92} & 0.09  \\
     \hline
     \multirow{3}{*}{16} & FT & 33.4 & 0.29 & 0.57 & 0.63 & 0.92 & 0.14 \\
     & PT & 33.9 & 0.29 & 0.57 & 0.63 & 0.92 & 0.15 \\
     & SPT & 34.1 & \textbf{0.30} & 0.55 & \textbf{0.64} & \textbf{0.93} & \textbf{0.17} \\
     & Com & \textbf{35.9} &  0.29 & \textbf{0.56} & 0.63 & \textbf{0.93} & 0.14  \\
     \hline
     \multirow{4}{*}{50} & FT & 36.4 & 0.33 & 0.57 & \textbf{0.65} & 0.93 & 0.24 \\
     & PT & 33.7 & 0.30 & 0.55 & 0.64 & 0.93 & 0.17 \\
     & SPT & 35.2 & 0.31	& 0.54 & \textbf{0.65} & 0.93 & 0.21 \\
     & Com & \textbf{37.2}  & \textbf{0.34} & \textbf{0.52} & \textbf{0.66} & 0.93 & \textbf{0.27}  \\
     \hline
     \multirow{4}{*}{100} & FT & 39.2 & \textbf{0.34} & 0.53 & \textbf{0.66} & 0.93 & \textbf{0.30} \\
     & PT & 35.0 & 0.32 & 0.54 & 0.65 & 0.93 & 0.22 \\
     & SPT & 36.8 & 0.33 & 0.53 & 0.65 & 0.93 & 0.26  \\
     & Com & \textbf{39.8} & \textbf{0.34} & \textbf{0.52} & \textbf{0.66} & 0.93 & 0.29  \\
     \hline
     \multirow{4}{*}{All} & FT & 46.6 & 0.39 & 0.48 & 0.68 & 0.94 & 0.38 \\
     & PT & 46.0 & 0.38 & 0.47 & 0.68 & 0.94 & 0.39 \\
     & SPT & 46.7 & 0.39 & 0.47 & 0.68 & 0.94 & 0.40  \\
     & Adapter & \textbf{48.5} & \textbf{0.40} & \textbf{0.46} & \textbf{0.69} & 0.94 & \textbf{0.41}  \\
     \hline
    \end{tabular}
\end{center}
\caption{ Comprehensive evaluation results of several PEFT methods on DART in few-shot cases. FT denotes Fine-Tuning, PT represents Prompt-Tuning, and SPT denotes Scaled Prompt-Tuning. Com represents Compacter. MET denotes METEOR, MS represents MoverScore, and BS denotes BERTScore. All means the whole dataset is used during tuning. The top-performing PEFT method in each few-shot case is displayed for comparison. \label{petsome_dart}}
\end{table}

\subsubsection{Summary}

According to the comparison above, we draw several suggestions regarding the adoption of PEFT methods below. First, Compacter is always the candidate when a small number of instances are available considering its decent performance and modest tuning cost on all datasets. Second, the proposed SPT is the optimal selection when memory on device is limited while a large number of training instances are available. Third, Prefix-tuning and UniPELT are not good options in few-shot cases especially on challenging datasets.

\subsection{Prompt length}

The prompt length has an inevitable impact on the performance of SPT. We conduct SPT employing all the instances in each individual dataset to find the optimal settings, which are then used in few-shot scenarios. The configurations of varied PEFT methods in Tab.~\ref{pet_lr} are all determined from the all-shot case. Therefore, one argument is that the performance of aforementioned approaches could be further improved if the hyperparameter settings are especially designed for each few-shot case. This could be the case while the comparisons present above are still fair and reflect the robustness and generalization capability of the methods.

\begin{figure}[!th]
\centering
\begin{subfigure}[b]{0.4\textwidth}
    \includegraphics[width=\textwidth]{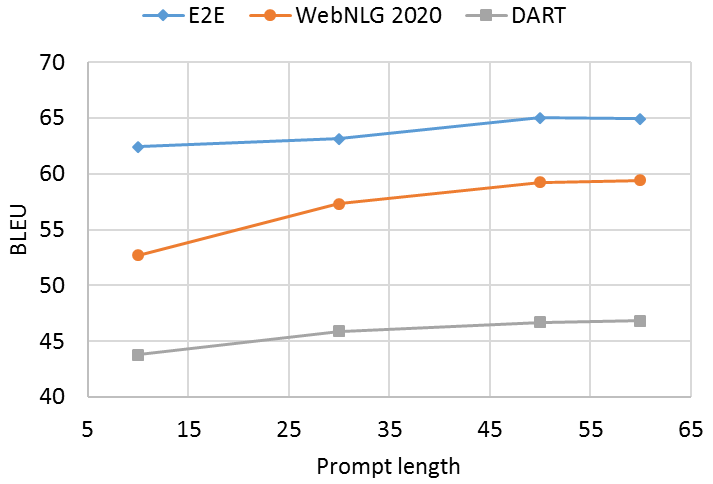}
    \caption{Prompt length vs. BLEU}
\end{subfigure}
\par\medskip
\begin{subfigure}[b]{0.4\textwidth}
    \includegraphics[width=\textwidth]{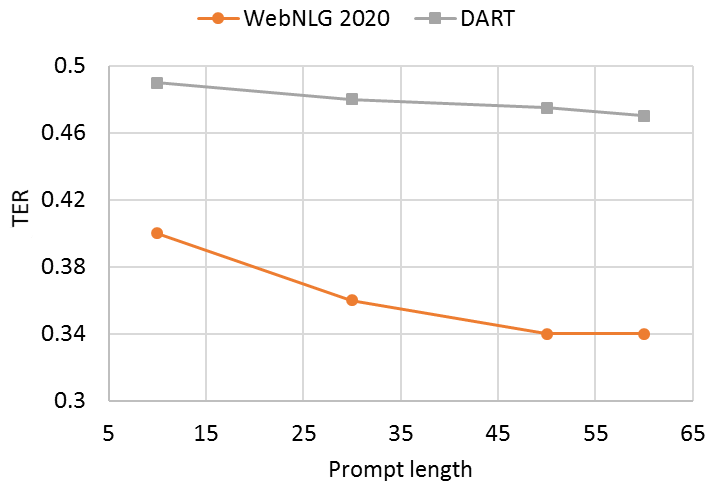}
    \caption{Prompt length vs. TER}
\end{subfigure}

\caption{The impact of prompt length on BLEU and TER for Scaled Prompt-Tuning. The results are obtained by whole-dataset tuning. The prompt lengths involved are 10, 30, 50, and 60. E2E dataset does not define any TER related evaluation metric.
\label{promptlen}}
\end{figure}

Fig.~\ref{promptlen} depicts prompt length vs. BLEU and TER on three datasets. BLEU scores raise up as the prompt becomes lengthier while the rate of growth gradually declines on three datasets. TER  decreases following the increase in prompt length on WebNLG 2020 and DART dataset. Thus, we set the prompt length to be 50, considering there is no significant performance gap  when the prompt length varies from 50 to 60.

\subsection{Tuning stability}

\begin{figure*}[!th]
\begin{subfigure}[b]{0.32\textwidth}
    \includegraphics[width=\textwidth]{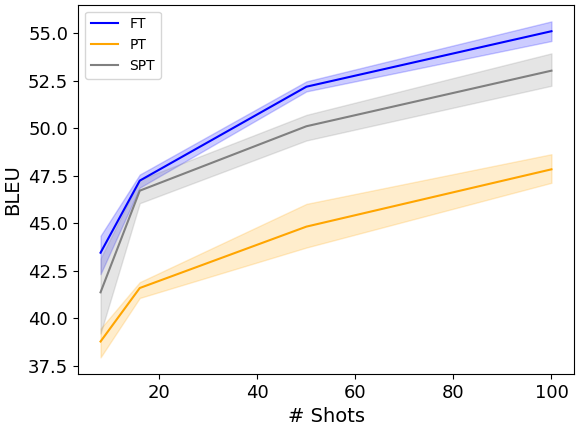}
    \caption{WebNLG 2020}
\end{subfigure}
\hfill
\begin{subfigure}[b]{0.32\textwidth}
    \includegraphics[width=\textwidth]{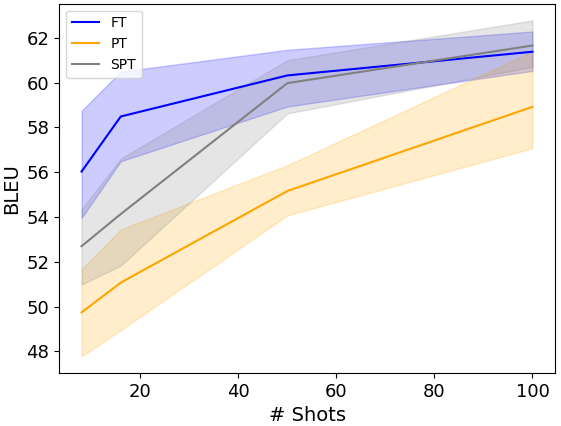}
    \caption{E2E}
\end{subfigure}
\hfill
\begin{subfigure}[b]{0.32\textwidth}
    \centering
    \includegraphics[width=\textwidth]{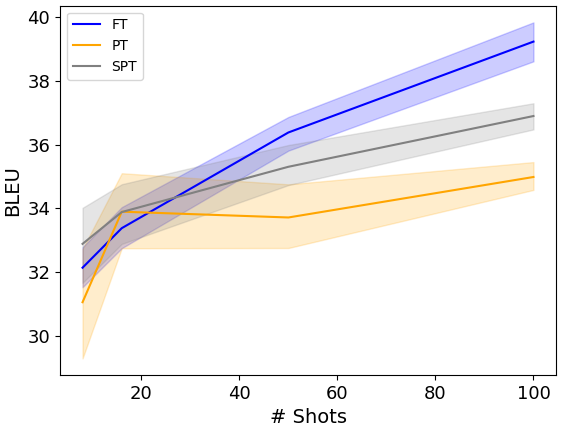}
    \caption{DART}
\end{subfigure}

\caption{The tuning stability of Fine-Tuning, Prompt-Tuning, and Scaled Prompt-Tuning on three datasets. \label{lineplot}}
\end{figure*}

According to previous study, fine-tuning on small downstream tasks is unstable. When it comes to few-shot PEFT, the stability issue could be severer. Fig.~\ref{lineplot} illustrates the average performance and error bands of three tuning methods in few-shot cases: Fine-Tuning, Prompt-Tuning, and SPT. Regarding datasets, tuning on WebNLG 2020 is more stable than others mainly because it has more instances involved in tuning under the same few-shot circumstance as others. Fine-Tuning is more stable than others, which is reasonable given its largest number of tunable parameters. Meanwhile, SPT is on-par with Fine-Tuning on E2E and obviously surpasses Prompt-Tuning on DART in terms of stability. Additionally, SPT demonstrates an almost absolute performance gain in comparison to Prompt-Tuning on WebNLG 2020 and E2E dataset.

\subsection{Multi-task Tuning vs. intermediate Tuning}

\begin{figure}[!th]
\centering
\begin{subfigure}[b]{0.45\textwidth}
    \centering
    \includegraphics[width=\textwidth]{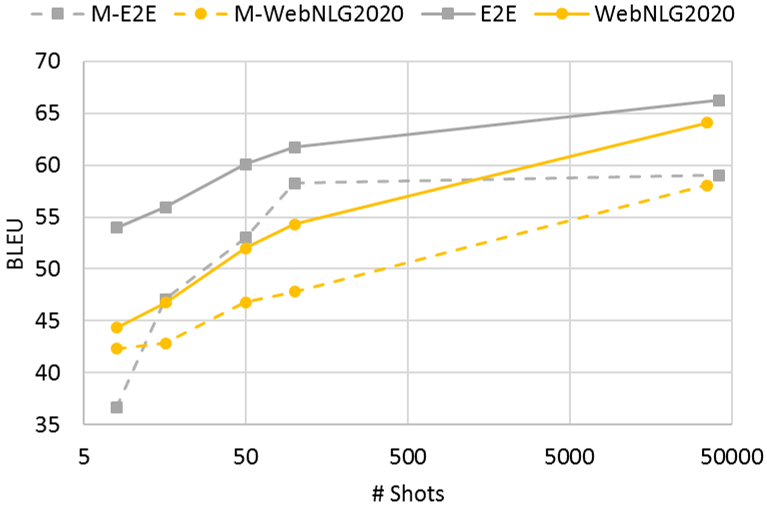}
    \caption{Fine-Tuning}
\end{subfigure}
\par\medskip
\begin{subfigure}[b]{0.45\textwidth}
    \centering
    \includegraphics[width=\textwidth]{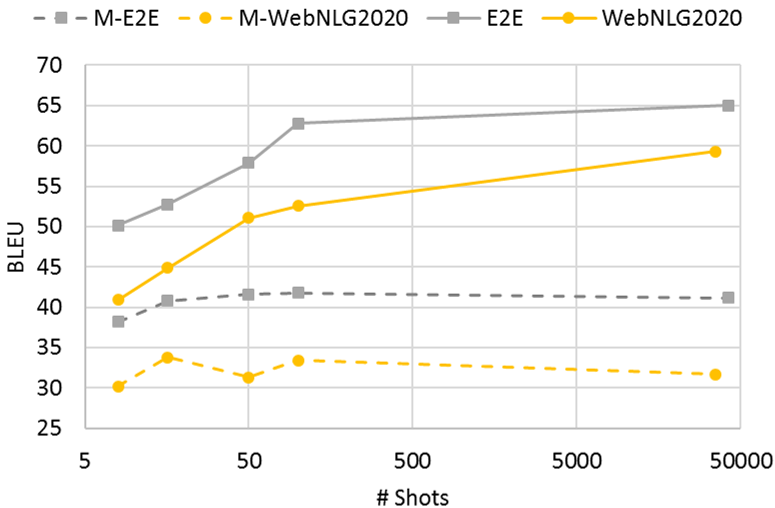}
    \caption{Scaled Prompt-Tuning}
\end{subfigure}
\caption{Multi-task Fine-tuning and Scaled Prompt-Tuning in few-shot cases. \emph{M-E2E} and \emph{M-WebNLG2020} denote evalution results on E2E and WebNLG 2020, respectively, after multi-task tuning. \emph{E2E} and \emph{WebNLG 2020} represent corresponding single-task tuning. \label{multi_tuning}}
\end{figure}

We further study how the proposed SPT performs in the paradigm of few-shot multi-task training and intermediate training, where WebNLG 2020 and E2E are involved. With respect to few-shot multi-task tuning, the same $n$-shot samples from two datasets, respectively, are mixed and used in tuning. According to Fig.~\ref{multi_tuning}, multi-task tuning always leads to the performance drop compared with single-task tuning, no matter which tuning method is present. As the number of shots rises, the performance of multi-task Fine-Tuning on two datasets gradually improves, while multi-task SPT presents a marginal improvement. The possible reason is that the two datasets are discrepant regarding data source and format, thus a single soft prompt is hard to reconcile them.

\begin{figure*}[!th]
\centering
\begin{subfigure}[b]{0.90\textwidth}
    \centering
    \includegraphics[width=\textwidth]{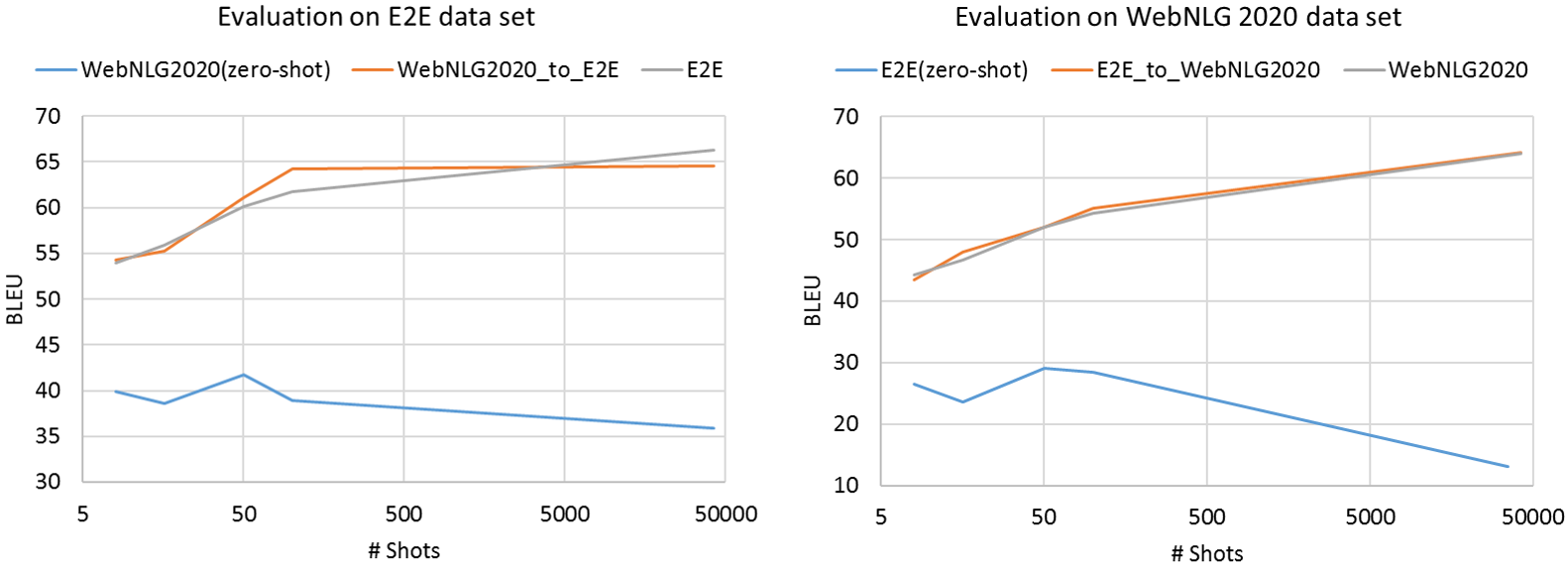}
    \caption{Intermediate Fine-Tuning}
\end{subfigure}
\par\medskip
\begin{subfigure}[b]{0.90\textwidth}
    \centering
    \includegraphics[width=\textwidth]{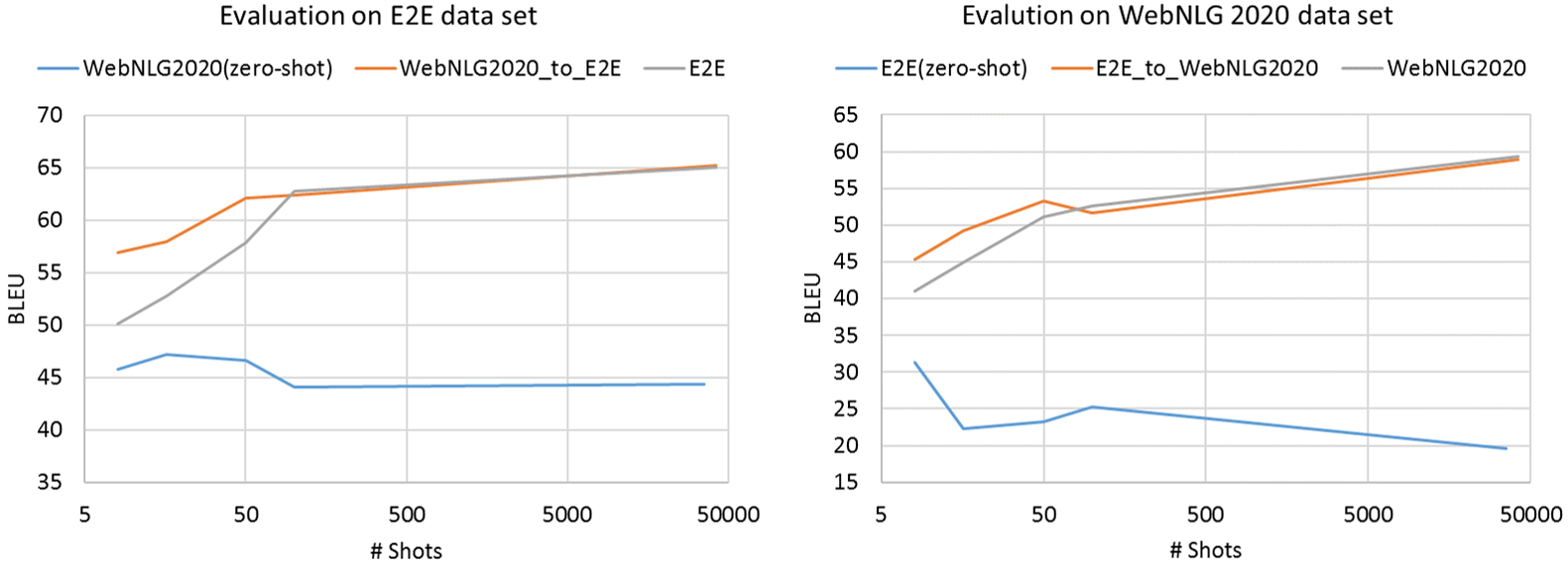}
    \caption{Intermediate Scaled Prompt-Tuning}
\end{subfigure}
\caption{The performance of intermediate Fine-Tuning and Scaled Prompt-Tuning in few-shot cases. Take the evaluation on E2E via intermediate Fine-Tuning in (a) as an example. \emph{E2E} represents using the model fine-tuned with E2E dataset for evaluation. \emph{WebNLG2020} represents that the model fine-tuned on WebNLG 2020 is evaluated on E2E, indicating the method's zero-shot capability. \emph{WebNLG2020\_to\_E2E} represents firstly fine-tuning on WebNLG 2020, then fine-tuning on E2E, and finally evaluating on E2E. \label{inter_tuning}}
\end{figure*}

Fig.~\ref{inter_tuning} applies the paradigm of intermediate training to Fine-Tuning and Scaled Prompt-Tuning to reveal their cross-task transferability. Two transferring directions are from WebNLG 2020 to E2E and from E2E to WebNLG 2020. According to Fig.~\ref{inter_tuning} (a), the model first fine-tuned on WebNLG 2020 and then fine-tuned on E2E outperforms the model merely fine-tuned on E2E in few-shot cases. However, the model first fine-tuned on E2E and then fine-tuned on WebNLG 2020 does not obviously surpasses the model only fine-tuned on WeNLG 2020. These manifest that Fine-Tuning results in the limited transferability of the parameters while the transfer direction is of significance. Intuitively, WebNLG 2020 is more complicated and challenging than E2E dataset, and the knowledge the parameters acquire after fine-tuned on WebNLG 2020 could be beneficial to E2E dataset, but not vice versa. This is the reason that we see the performance gain when the intermediate fine-tuning order is WebNLG 2020 first and E2E second. Additionally, we witness that the model fine-tuned on one dataset shows some zero-shot ability on the other dataset.

In terms of intermediate SPT, Fig.~\ref{inter_tuning} (b) elaborates a slightly different story. There are substantial performance enhancement in both transfer directions, from WebNLG 2020 to E2E and from E2E to WebNLG 2020, in few-shot cases. Moreover, SPT leads to the parameters' stronger zero-shot capability on E2E than Fine-Tuning. These provide the following suggestion in real-world applications. With a small number of available instances and limited computation resources in device, we could directly take the instances from existing datasets in similar task and conduct intermediate SPT. This does not necessitate additional labeling cost and large memory footprint while yielding decent performance.

\section{Conclusion}

In this work, we study few-shot PEFT methods on structured data-to-text generation tasks. Scaled Prompt-Tuning is proposed which almost does not introduce extra trainable parameters and computations than conventional Prompt-Tuning while drastically boosts the performance. Moreover, we comprehensively evaluate several PEFT methods in few-shot cases on three NLG datasets. Experiments demonstrate that no single approach could always prevail over others under all circumstances with a relatively small number of trainable parameters. Meanwhile, the performance of certain methods such as Prefix-Tuning and UniPELT could dramatically deteriorate in few-shot cases on challenging datasets such as DART.
We further study multi-task tuning and intermediate tuning combined with PEFT methods under few-shot circumstances. The proposed SPT approach showcases a decent transferability in few-shot cases. This provides a promising scheme in scenarios where a limited number of instances from downstream tasks are available, which does not introduce any extra labeling cost or require large memory footprint.

\bibliography{bibliography}

\begin{thebibliography}{22}
\expandafter\ifx\csname natexlab\endcsname\relax\def\natexlab#1{#1}\fi

\bibitem[{Aghajanyan et~al.(2020)Aghajanyan, Zettlemoyer, and
  Gupta}]{aghajanyan2020intrinsic}
Armen Aghajanyan, Luke Zettlemoyer, and Sonal Gupta. 2020.
\newblock Intrinsic dimensionality explains the effectiveness of language model
  fine-tuning.
\newblock \emph{arXiv preprint arXiv:2012.13255}.

\bibitem[{Brown et~al.(2020)Brown, Mann, Ryder, Subbiah, Kaplan, Dhariwal,
  Neelakantan, Shyam, Sastry, Askell et~al.}]{brown2020language}
Tom Brown, Benjamin Mann, Nick Ryder, Melanie Subbiah, Jared~D Kaplan, Prafulla
  Dhariwal, Arvind Neelakantan, Pranav Shyam, Girish Sastry, Amanda Askell,
  et~al. 2020.
\newblock Language models are few-shot learners.
\newblock \emph{Advances in neural information processing systems},
  33:1877--1901.

\bibitem[{Devlin et~al.(2018)Devlin, Chang, Lee, and
  Toutanova}]{devlin2018bert}
Jacob Devlin, Ming-Wei Chang, Kenton Lee, and Kristina Toutanova. 2018.
\newblock Bert: Pre-training of deep bidirectional transformers for language
  understanding.
\newblock \emph{arXiv preprint arXiv:1810.04805}.

\bibitem[{Hambardzumyan et~al.(2021)Hambardzumyan, Khachatrian, and
  May}]{hambardzumyan2021warp}
Karen Hambardzumyan, Hrant Khachatrian, and Jonathan May. 2021.
\newblock Warp: Word-level adversarial reprogramming.
\newblock \emph{arXiv preprint arXiv:2101.00121}.

\bibitem[{He et~al.(2021)He, Zhou, Ma, Berg-Kirkpatrick, and
  Neubig}]{he2021towards}
Junxian He, Chunting Zhou, Xuezhe Ma, Taylor Berg-Kirkpatrick, and Graham
  Neubig. 2021.
\newblock Towards a unified view of parameter-efficient transfer learning.
\newblock \emph{arXiv preprint arXiv:2110.04366}.

\bibitem[{Houlsby et~al.(2019)Houlsby, Giurgiu, Jastrzebski, Morrone,
  De~Laroussilhe, Gesmundo, Attariyan, and Gelly}]{houlsby2019parameter}
Neil Houlsby, Andrei Giurgiu, Stanislaw Jastrzebski, Bruna Morrone, Quentin
  De~Laroussilhe, Andrea Gesmundo, Mona Attariyan, and Sylvain Gelly. 2019.
\newblock Parameter-efficient transfer learning for nlp.
\newblock In \emph{International Conference on Machine Learning}, pages
  2790--2799. PMLR.

\bibitem[{Hu et~al.(2021)Hu, Shen, Wallis, Allen-Zhu, Li, Wang, Wang, and
  Chen}]{hu2021lora}
Edward~J Hu, Yelong Shen, Phillip Wallis, Zeyuan Allen-Zhu, Yuanzhi Li, Shean
  Wang, Lu~Wang, and Weizhu Chen. 2021.
\newblock Lora: Low-rank adaptation of large language models.
\newblock \emph{arXiv preprint arXiv:2106.09685}.

\bibitem[{Karimi~Mahabadi et~al.(2021)Karimi~Mahabadi, Henderson, and
  Ruder}]{karimi2021compacter}
Rabeeh Karimi~Mahabadi, James Henderson, and Sebastian Ruder. 2021.
\newblock Compacter: Efficient low-rank hypercomplex adapter layers.
\newblock \emph{Advances in Neural Information Processing Systems},
  34:1022--1035.

\bibitem[{Lester et~al.(2021)Lester, Al-Rfou, and Constant}]{lester2021power}
Brian Lester, Rami Al-Rfou, and Noah Constant. 2021.
\newblock The power of scale for parameter-efficient prompt tuning.
\newblock \emph{arXiv preprint arXiv:2104.08691}.

\bibitem[{Lewis et~al.(2019)Lewis, Liu, Goyal, Ghazvininejad, Mohamed, Levy,
  Stoyanov, and Zettlemoyer}]{lewis2019bart}
Mike Lewis, Yinhan Liu, Naman Goyal, Marjan Ghazvininejad, Abdelrahman Mohamed,
  Omer Levy, Ves Stoyanov, and Luke Zettlemoyer. 2019.
\newblock Bart: Denoising sequence-to-sequence pre-training for natural
  language generation, translation, and comprehension.
\newblock \emph{arXiv preprint arXiv:1910.13461}.

\bibitem[{Li and Liang(2021)}]{li2021prefix}
Xiang~Lisa Li and Percy Liang. 2021.
\newblock Prefix-tuning: Optimizing continuous prompts for generation.
\newblock \emph{arXiv preprint arXiv:2101.00190}.

\bibitem[{Liu et~al.(2022)Liu, Tam, Muqeeth, Mohta, Huang, Bansal, and
  Raffel}]{liu2022few}
Haokun Liu, Derek Tam, Mohammed Muqeeth, Jay Mohta, Tenghao Huang, Mohit
  Bansal, and Colin~A Raffel. 2022.
\newblock Few-shot parameter-efficient fine-tuning is better and cheaper than
  in-context learning.
\newblock \emph{Advances in Neural Information Processing Systems},
  35:1950--1965.

\bibitem[{Liu et~al.(2021)Liu, Ji, Fu, Tam, Du, Yang, and Tang}]{liu2021p}
Xiao Liu, Kaixuan Ji, Yicheng Fu, Weng~Lam Tam, Zhengxiao Du, Zhilin Yang, and
  Jie Tang. 2021.
\newblock P-tuning v2: Prompt tuning can be comparable to fine-tuning
  universally across scales and tasks.
\newblock \emph{arXiv preprint arXiv:2110.07602}.

\bibitem[{Mao et~al.(2021)Mao, Mathias, Hou, Almahairi, Ma, Han, Yih, and
  Khabsa}]{mao2021unipelt}
Yuning Mao, Lambert Mathias, Rui Hou, Amjad Almahairi, Hao Ma, Jiawei Han,
  Wen-tau Yih, and Madian Khabsa. 2021.
\newblock Unipelt: A unified framework for parameter-efficient language model
  tuning.
\newblock \emph{arXiv preprint arXiv:2110.07577}.

\bibitem[{Pfeiffer et~al.(2020)Pfeiffer, Kamath, R{\"u}ckl{\'e}, Cho, and
  Gurevych}]{pfeiffer2020adapterfusion}
Jonas Pfeiffer, Aishwarya Kamath, Andreas R{\"u}ckl{\'e}, Kyunghyun Cho, and
  Iryna Gurevych. 2020.
\newblock Adapterfusion: Non-destructive task composition for transfer
  learning.
\newblock \emph{arXiv preprint arXiv:2005.00247}.

\bibitem[{Raffel et~al.(2020)Raffel, Shazeer, Roberts, Lee, Narang, Matena,
  Zhou, Li, and Liu}]{raffel2020exploring}
Colin Raffel, Noam Shazeer, Adam Roberts, Katherine Lee, Sharan Narang, Michael
  Matena, Yanqi Zhou, Wei Li, and Peter~J Liu. 2020.
\newblock Exploring the limits of transfer learning with a unified text-to-text
  transformer.
\newblock \emph{The Journal of Machine Learning Research}, 21(1):5485--5551.

\bibitem[{R{\"u}ckl{\'e} et~al.(2020)R{\"u}ckl{\'e}, Geigle, Glockner, Beck,
  Pfeiffer, Reimers, and Gurevych}]{ruckle2020adapterdrop}
Andreas R{\"u}ckl{\'e}, Gregor Geigle, Max Glockner, Tilman Beck, Jonas
  Pfeiffer, Nils Reimers, and Iryna Gurevych. 2020.
\newblock Adapterdrop: On the efficiency of adapters in transformers.
\newblock \emph{arXiv preprint arXiv:2010.11918}.

\bibitem[{Sanh et~al.(2021)Sanh, Webson, Raffel, Bach, Sutawika, Alyafeai,
  Chaffin, Stiegler, Scao, Raja et~al.}]{sanh2021multitask}
Victor Sanh, Albert Webson, Colin Raffel, Stephen~H Bach, Lintang Sutawika,
  Zaid Alyafeai, Antoine Chaffin, Arnaud Stiegler, Teven~Le Scao, Arun Raja,
  et~al. 2021.
\newblock Multitask prompted training enables zero-shot task generalization.
\newblock \emph{arXiv preprint arXiv:2110.08207}.

\bibitem[{Sung et~al.(2021)Sung, Nair, and Raffel}]{sung2021training}
Yi-Lin Sung, Varun Nair, and Colin~A Raffel. 2021.
\newblock Training neural networks with fixed sparse masks.
\newblock \emph{Advances in Neural Information Processing Systems},
  34:24193--24205.

\bibitem[{Wang et~al.(2018)Wang, Singh, Michael, Hill, Levy, and
  Bowman}]{wang2018glue}
Alex Wang, Amanpreet Singh, Julian Michael, Felix Hill, Omer Levy, and Samuel~R
  Bowman. 2018.
\newblock Glue: A multi-task benchmark and analysis platform for natural
  language understanding.
\newblock \emph{arXiv preprint arXiv:1804.07461}.

\bibitem[{Zaken et~al.(2021)Zaken, Ravfogel, and Goldberg}]{zaken2021bitfit}
Elad~Ben Zaken, Shauli Ravfogel, and Yoav Goldberg. 2021.
\newblock Bitfit: Simple parameter-efficient fine-tuning for transformer-based
  masked language-models.
\newblock \emph{arXiv preprint arXiv:2106.10199}.

\bibitem[{Zhong et~al.(2021)Zhong, Friedman, and Chen}]{zhong2021factual}
Zexuan Zhong, Dan Friedman, and Danqi Chen. 2021.
\newblock Factual probing is [mask]: Learning vs. learning to recall.
\newblock \emph{arXiv preprint arXiv:2104.05240}.

\end{thebibliography}
\bibliographystyle{acl_natbib}

\end{document}